\documentclass[letterpaper, 10 pt, conference]{ieeeconf}  

\IEEEoverridecommandlockouts                              

\usepackage{xcolor}         
\usepackage{graphicx}
\usepackage{caption}
\usepackage{amsmath}
\usepackage{amssymb}
\usepackage{amsfonts}
\usepackage{booktabs}
\usepackage{multirow}
\usepackage{tabularx}
\usepackage{url}

\title{\LARGE \bf
Autonomous Robotic Assembly: From Part Singulation to \\Precise Assembly
}

\author{Kei Ota$^{1,2}$,  Devesh K. Jha$^{1,2,3}$, Siddarth Jain$^{1}$, Bill Yerazunis$^{1}$, Radu Corcodel$^{1}$, Yash Shukla$^{1}$,\\ Antonia Bronars$^{1}$, Diego Romeres$^{1}$
}

\begin{document}
 \twocolumn[{%
    \renewcommand\twocolumn[1][]{#1}%
    \maketitle
    \begin{center}
        \centering
        \vspace{-5mm}
        \includegraphics[width=1.0\textwidth]{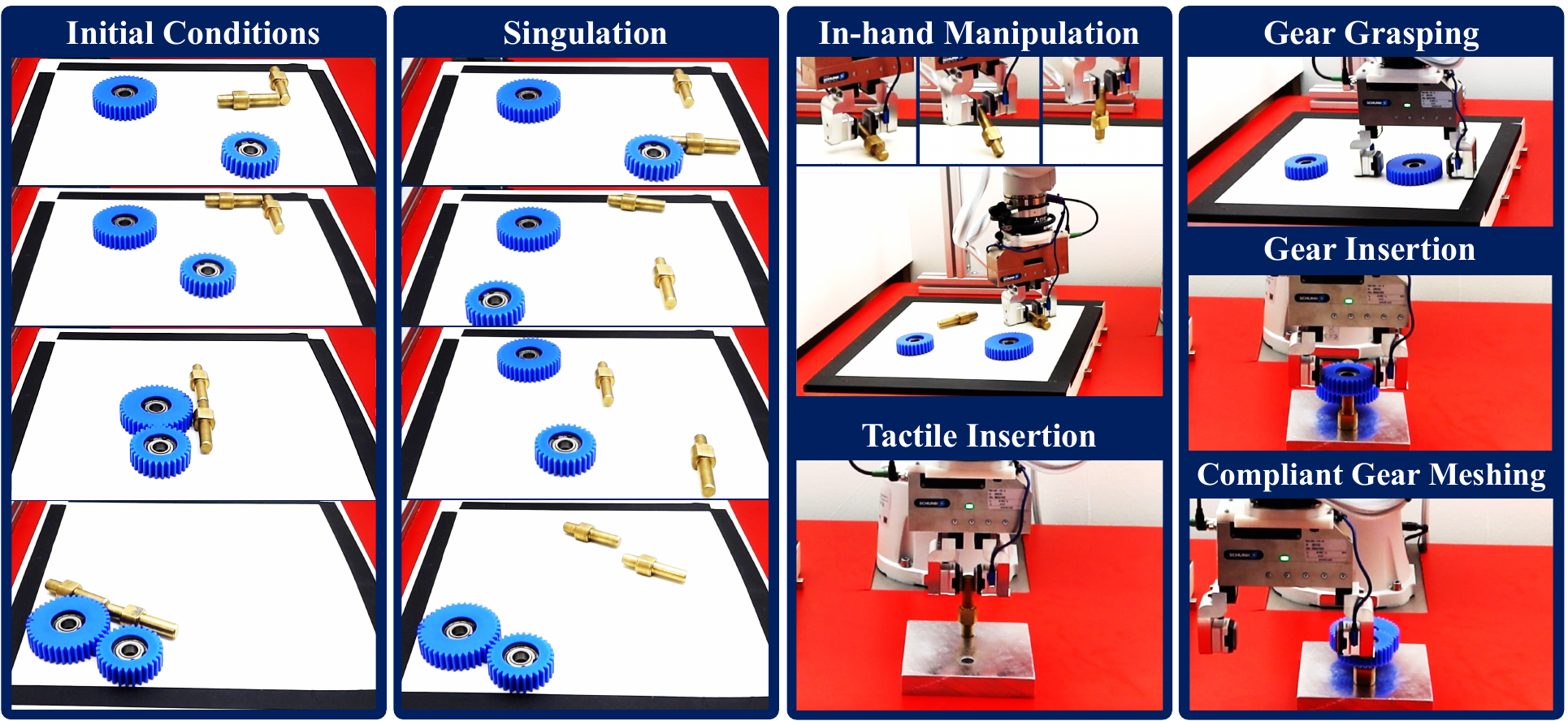}
        \captionof{figure}{\textbf{Autonomous Robotic Assembly:} We present an autonomous robotic assembly system that can assemble a gear box from any given initial condition, as shown in the first column. The assembly system can reason about grasp feasibility and slide selected objects out of a clutter to create grasp affordances for the assembly parts. Then it performs the pose manipulation and grasping required for the downstream assembly task. Finally, using various different controllers, it performs the required insertion and meshing of gears to assemble a functioning gear box. The proposed system works in a closed-loop fashion, where it can deliberate on the success and failure of individual steps and react accordingly. 
        }
        \label{fig:intro_fig}
    \end{center}%
    }]
\footnotetext[1]{All authors are with Mitsubishi Electric Research Labs (MERL), Cambridge, MA 02139}
\footnotetext[2]{Authors with equal contribution.}
\footnotetext[3]{Corresponding author. {\tt\small jha@merl.com}}

\begin{abstract}

Imagine a robot that can assemble a functional product from the individual parts presented in any configuration to the robot. Designing such a robotic system is a complex problem which presents several open challenges. To bypass these challenges, the current generation of assembly systems is built with a lot of system integration effort to provide the structure and precision necessary for assembly. These systems are mostly responsible for part singulation, part kitting, and part detection, which is accomplished by intelligent system design. In this paper, we present autonomous assembly of a gear box with minimum requirements on structure. The assembly parts are randomly placed in a two-dimensional work environment for the robot. The proposed system makes use of several different manipulation skills such as sliding for grasping, in-hand manipulation, and insertion to assemble the gear box. All these tasks are run in a closed-loop fashion using vision, tactile, and Force-Torque (F/T) sensors. We perform extensive hardware experiments to show the robustness of the proposed methods as well as the overall system.
See supplementary video at \url{https://www.youtube.com/watch?v=cZ9M1DQ23OI}.
\end{abstract}

\section{Introduction}\label{sec:introduction}
Designing robots that can achieve human-like dexterity and reasoning while performing complex, long-horizon tasks has been the long-standing goal of robotics. While robots have been getting very good at performing repetitive pick-and-place kind of operations, achieving reliable \& high degrees of dexterity remains elusive.
Consequently, robots require a lot of structure to perform tasks that require complex and long-horizon reasoning.
Imagine a factory floor where assembly blueprints and parts are presented to a robot in arbitrary configurations, and the robot can autonomously assemble the desired product~\cite{zhu2023multi}. While this task is effortless for humans, it is quite challenging for robotic systems. In this paper, we study a simplified assembly problem to understand the challenges of the underlying long-horizon manipulation. 

Assembly is arguably the single biggest application of robots in current society. However, this generally requires creating a complex system for singulation and manipulating the pose of parts. Robots are simply used as machines to perform a very precise pick-and-place task. Robots with intelligent manipulation skills can simplify the design and operation of these systems on factory floors. However,  manipulation is a complex problem from the aspects of planning, sensing, and control. Despite being an active area of research for decades, it still remains a widely open and challenging problem to all of robotics community~\cite{mason2018toward}. Through a simplified example of the assembly of a gear box (see Fig.~\ref{fig:intro_fig}), we highlight the challenges associated with designing agents that can achieve high levels of dexterity and autonomy with simple hardware.

  

Robotic assembly is a challenging, long-horizon manipulation problem. Thus, performing autonomous assembly requires reasoning about various sub-goals given a particular instance of the environment. Apart from reasoning about a feasible plan, it also presents the challenge of achieving very high accuracy and precision to make a tight tolerance assembly feasible. Another major challenge is to allow robustness to various initial conditions so that the system can always achieve the desired goal regardless of the initial condition. To demonstrate all these challenges, we present the assembly of a gear box shown in Fig.~\ref{fig:intro_fig}. Under some mildly restrictive assumptions, we present solutions to the problems in order to create an autonomous agent that can perform assembly with a perfect success rate. We make use of three different kinds of sensors -- an RGBD camera, vision-based Gelsight tactile sensors, and a six-axis F/T sensor mounted on the wrist of the manipulator. 

As shown in Fig.~\ref{fig:intro_fig}, our system can perform part singulation, part manipulation, as well as assembly in a closed-loop fashion while using an array of controllers and perception methods for individual steps. In particular, this paper has the following contributions:
\begin{enumerate}
    \item We present a benchmark assembly task which can be used to test the dexterity and precision of related manipulation techniques.
    \item We present an autonomous assembly system which can combine various sensing modalities such as vision, touch, and F/T while running in closed-loop feedback. The proposed system is extensively tested on hardware and has been shown to achieve an almost perfect success rate.
\end{enumerate}
To the best of our knowledge, an autonomous, multi-modal system that can perform high-precision multistep assembly task has not been presented earlier in the literature. We hope that the proposed system can be used as a benchmark in the near future for testing manipulation algorithms and methods. 


\section{Related Work}\label{sec:related_work}
\textbf{Robotic manipulation primitives.}
Autonomous assembly has been one of the most challenging and widely studied problems in robotics in the past several decades~\cite{knepper2013ikeabot,thomas2018learning, suarez2018can,kimbel2020benchmarking}. Several challenges related to assembly tasks have been widely studied in the literature, including grasping in clutter~\cite{mahler2019learning, kiatos2019robust, lundell2021ddgc}, part singulation~\cite{zeng2018learning}, pose estimation~\cite{xiang2017posecnn, tremblay2018deep, bauza2023tac2pose, ota2023tactile}, part insertion~\cite{9561646}, etc. 
Similarly, in-hand manipulation and non-prehensile manipulation have also been widely studied in the literature~\cite{zhang2023simultaneous, 9811812, shirai2023robust, chavan2020planar, kim2024texterity, liang2024robust}. However, all these work would require an external point of contact for manipulation.
In-hand pose estimation using tactile sensors has also been explored recently~\cite{bauza2023tac2pose, ota2023tactile}. Similarly, insertion using force or tactile feedback has also been explored~\cite{9561646, 10178229}. However, these methods could struggle for cases with very tight tolerances required during assembly.
Furthermore, these primitive skills have not been composed for long horizon tasks with sensory feedback.

\textbf{Robotic assembly benchmarks.}
Long-horizon manipulation involves many challenges, including the need to dynamically switch skills based on observations and the ability to recover from failures. Recent robotic benchmarks, such as furniture assembly~\cite{knepper2013ikeabot,suarez2018can,heo2023furniturebench}, irregular block stacking~\cite{lee2021rgbstacking}, and manufacturing tasks~\cite{suarez2018can,kimbel2020benchmarking,heo2023furniturebench}, provide reproducible environments to evaluate the manipulation performance. 
Our task is inspired by the NIST robotic manipulation benchmark~\cite{kimbel2020benchmarking}, employing a simplified insertion tolerance of approximately 1 mm as opposed to the submillimeter tolerance ($0.029$ mm and $0.005$ mm) of the standard board. This simplication was explored similarly in the NIST-i benchmark, but focused only on gear insertion~\cite{bousmalis2024robocat}. 
Previous approaches to solve these assembly benchmarks typically omit tactile feedback, relying heavily on vision sensors, leading to issues with occlusion or kinematic constraints imposed by extra links for vision sensors mounted on robot links. Although tactile feedback has been explored for pick-and-place tasks~\cite{bauza2023simple}, we believe that its integration with diverse visuotactile techniques for the design of autonomous assembly systems represents a significant and novel contribution to the existing literature.

\section{Problem Statement}
While the autonomous assembly problem in its entirety is a very complex problem, we make certain assumptions to solve it in a limited scope. This section briefly describes it.

\subsection{Task Assumptions}
The main objective of the paper is to evaluate the manipulation capability and robustness of the proposed system. In order to evaluate these, we make some simplifying assumptions to limit the scope of this study.
\begin{enumerate}
    \item The positions of the holes are known and fixed.
    \item The surface for the manipulation of parts and the base plate for assembly is perfectly leveled (see Fig.~\ref{fig:intro_fig}).
    \item The assembly order of the parts is known.
\end{enumerate}
Assumption (1) is not very restrictive. We can easily relax this by using a vision system to track holes, as shown earlier in our previous work~\cite{10178229}. The tray is kept leveled so that the parts do not slide under gravity. However, in the most general case, the manipulation primitives can be generalized to an inclined one. Assumption (3) can also be relaxed by using a Large Language Model~\cite{sun2023interactive} or a graph planner~\cite{zhu2023multi}. 
\subsection{System Description} \label{subsec:system_description}
\textbf{Robot platform.}
The MELFA RV-5AS-D Assista robot, a 6 DoF collaborative robot, is used in this study. Two tactile sensors are mounted on the WSG-50 gripper, and a Force-Torque (F/T) sensor is mounted on the wrist (see the supplementary video for the system). The robot fingers also have a silicone gel padding at the bottom, which is used during contact interactions.

\textbf{Tactile sensor.}
We use commercially available GelSight Mini tactile sensors~\cite{gelsightmini2023}, which provides 320×240 compressed RGB images at a rate of approximately 25 Hz, with a field of view of 18.6 × 14.3 millimeters.

\textbf{Vision sensor.}
An Intel Realsense L515 camera is used as a vision sensor for the RGB-D input of the scene. This vision sensor is used to observe the table-top used for manipulation of assembly parts. The assembly location is known to the robot and is not observed by the vision sensor.

\begin{figure}
    \centering
    \includegraphics[width=\columnwidth]{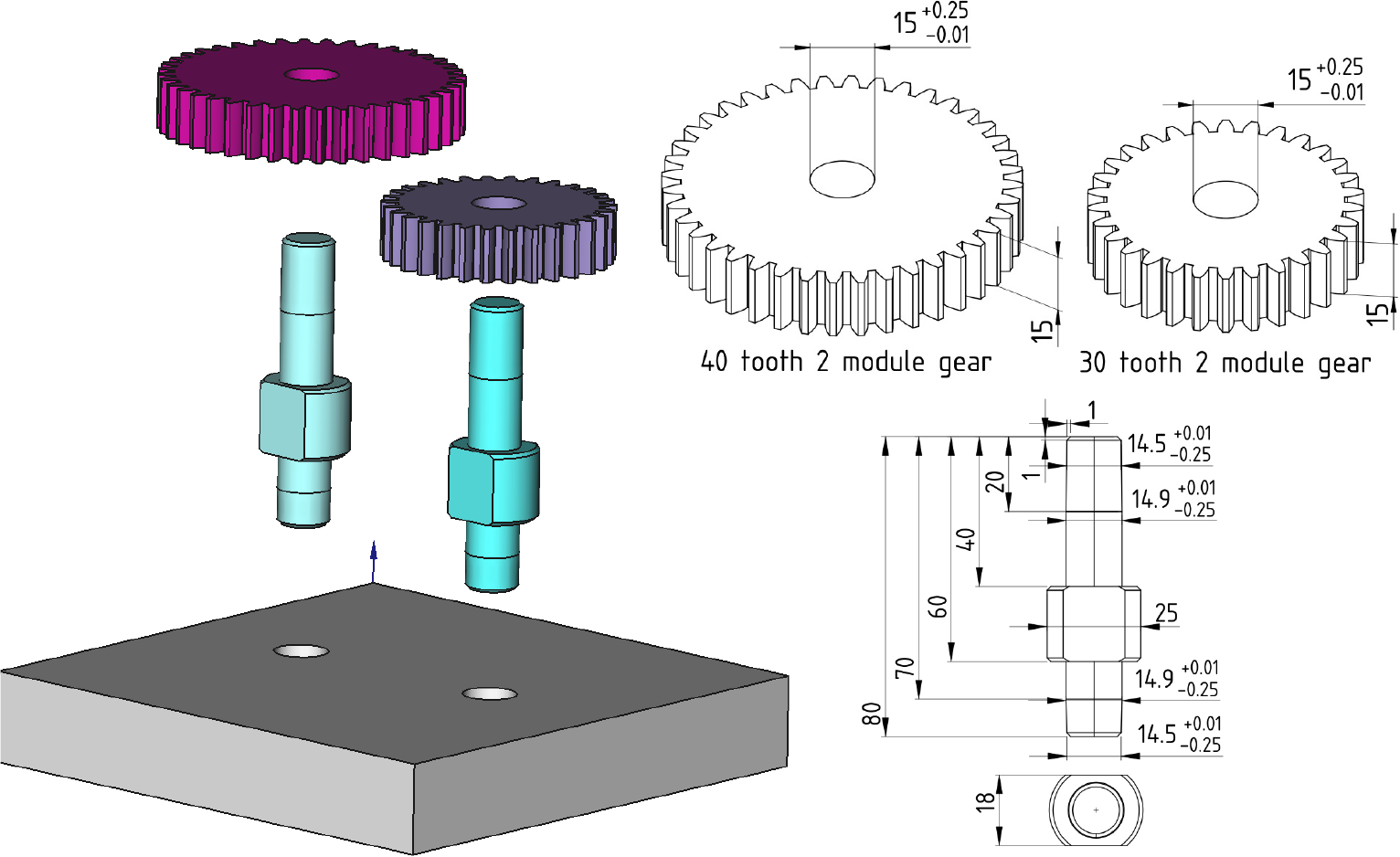}
    \caption{This figure shows the assembly parts with two identical pegs and two gears. Accurate dimensions of the parts are provided up to machining tolerances. The holes in the base plate are 15 mm in diameter and $70$ mm apart.
    }
    \label{fig:assembly_parts}
\end{figure}

\textbf{Assembly Parts.}
Figure~\ref{fig:assembly_parts} shows the assembly parts. The gear box consists of two pegs, a small and a large gear, and a baseplate that has two holes for the pegs. The dimensions of the individual parts are shown in Figure~\ref{fig:assembly_parts}. The tolerance for the various insertion tasks is approximately $1$ mm. While the tolerances are much higher than what is expected in assembled products~\cite{kimbel2020benchmarking}, we believe the proposed approaches could be used for tighter assembly.  We denote the parts by $p_1,p_2,g_l,g_s$ for the pegs, the large gear, and the small gear, respectively. 

\textbf{Assembly Task.} Imagine a task where assembly parts are presented in any given configuration in the robot workspace. The robot has to assemble a functional gear box by manipulating these parts. As shown in the first column of Fig.~\ref{fig:intro_fig}, we present the parts in any planar configuration to the robot. The task for the robot is to perform the required manipulation and assemble a functioning gear box using these parts. It is noted that the parts are always presented in a workspace that is visible to an overhead camera (the black boundaries in Fig.~\ref{fig:intro_fig} show this workspace). As specified earlier, the location of the base plate where the assembly has to be performed is known to the robot.

\textbf{What makes this task challenging?} There are multiple tasks that an agent needs to reason about for the assembly task. The first task is to reason about grasp affordances when the parts are presented in a clutter. In case the parts cannot be grasped directly, the robot has to reason about how it can move them apart or singulate them for grasping. The second challenge is for the robot to reason about how it should grasp the individual parts so that it is suitable for the desired downstream assembly task. Consequently, how should the robot achieve the precision required for assembly when the grasp pose is not fixed? Furthermore, how can the system be designed to achieve extremely robust performance? We address some of these challenges in this paper.

\section{System Details}\label{sec:system}
\begin{figure*}[t]
    \centering
    \includegraphics[width=0.9\textwidth]{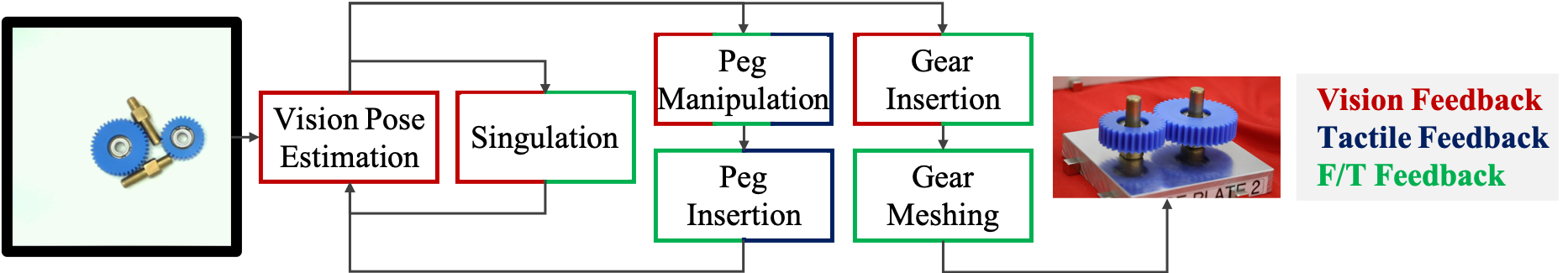}
    \caption{System level overview of the assembly controller. The color codes indicate the feedback modality for the particular related operation. Multiple color blocks for the same operation indicate that multiple sensors are used for feedback and/or controller design for the particular operation. [Best seen in color].
    }
    \label{fig:pipeline}
\vspace{-0.2cm}    
\end{figure*}

A system-level flow chart of the designed assembly system is shown in Fig.~\ref{fig:pipeline}. As could be seen, most steps use multi-modal feedback during execution of the task. We believe that this provides robustness to the system to perform the task reliably.

\subsection{Vision System}
Achieving reliable robotic grasping hinges on accurately determining the pose of target objects within the input data. Recent advancements in deep learning-based object pose estimation~\cite{xiang2017posecnn, tremblay2018deep, wen2023bundlesdf} have garnered attention for their ability to achieve centimeter-level accuracy across diverse datasets. Nevertheless, these methods are often insufficient for meeting the stringent requirements of precise robotic manipulation tasks at the sub-millimeter level. Moreover, in our scenario, assembly parts are initially placed in arbitrary configurations, involving multiple instances of the same object type. 

In our settings, the workspace of the robot for autonomous assembly is constrained in that the parts lie on a planar workspace, and thus, the grasp synthesis can be constrained from one direction. Therefore, the synthesis of the grasp of each object can define a pose output, $g = (p, \phi, q)$, where $p = (x, y, z$) is the position of the object, $\phi$ is the singular rotation angle around the vertical axis, and $q$ represents the label of the object class. We adopt a hybrid approach that involves deep learning for part detection and segmentation, combined with analytical pose computation. The network architecture is built on a backbone convolutional neural network architecture for feature extraction. We use a feature pyramid network (FPN) based on ResNet-50~\cite{he2016deep}. 
Mask-RCNN~\cite{he2017mask} is based on region proposals that are generated through a region proposal network. 
It adds a network head, a fully convolutional neural network, to produce the desired instance segmentation. Finally, mask and class predictions are decoupled; the mask network head predicts the mask independently from the network head predicting the class. Typically, this involves using a multitask loss function $L$ = $L_{cls} + L_{bbox} + L_{mask}$. For training the network, we perform transfer learning from the MS COCO dataset pre-trained weights in a supervised manner. We capture $380$ images of size $640\times 480$ under different initial conditions and annotate the data to indicate segmentation pixels and class labels. A 5:1 ratio was used to divide the dataset into training and validation sets. We selected the maximum iterations of 4000. We identify the resulting segmentation masks for parts with the network prediction. The detected segmentation masks are utilized with depth estimation to compute the corresponding registered point cloud data points. These clusters are then processed to perform analytic pose estimation~\cite{jain2016grasp} for grasping and part singulation. We analyze the principal components of a covariance matrix created from the nearest neighbours of the point $\boldsymbol{p}_i$ in a local neighbourhood of size $n$. We denote the centroid $\bar{\boldsymbol{p}}$ of the neighbours of $\boldsymbol{p}_i$, given by Eq.~\eqref{eq:position}. The covariance matrix $C \in \mathbb{R}^{3\times3}$ is computed as in Eq.~\eqref{eq:Cmat} and the eigenvector problem is presented in Eq.~\eqref{eq:eigen}. The eigenvalues $\lambda_j \in \mathbb{R}$ and the eigenvectors $\vec{v}_j$ form an orthogonal frame, corresponding to the principal components of the point set in the neighbourhood $n$.

\begin{equation}
\boldsymbol{ \bar p} = \frac{\sum_{k=1}^{n} \boldsymbol{p}_k}{n},    \label{eq:position}
\end{equation}
\begin{equation}
C = \frac{1}{n} \sum_{i=1}^{n} (\boldsymbol{p}_i - \boldsymbol{ \bar p}) (\boldsymbol{p}_i - \boldsymbol{ \bar p})^T ,  
\label{eq:Cmat}
\end{equation}
\begin{equation}
C \cdot \vec v_l = \lambda_l \cdot \vec v_l, \, l \in \{0, 1, 2 \}.
\label{eq:eigen}
\end{equation}

The eigenvalue $\lambda_l$ quantifies the variation of $\boldsymbol{p}_i$ along the direction of the corresponding eigenvector and the smallest eigenvalue $\lambda_0  (\lambda_0 \leqslant \lambda_1 \leqslant \lambda_2)$ corresponds to the variation along the surface normal vector (i.e. the eigenvector $\vec v_0)$ of the surface patch of neighbourhood size $n$. The object pose is then represented 
with $\bar{\boldsymbol{p}}$ and the singular rotation angle around the vertical axis $\phi$ computed from the angle between the minor eigenvector and the horizontal
axis. The object class label $q$ is obtained from the Mask-RCNN output. 

\begin{figure*}[t]
\begin{minipage}{0.63\linewidth}
    \centering
    \includegraphics[width=\columnwidth]{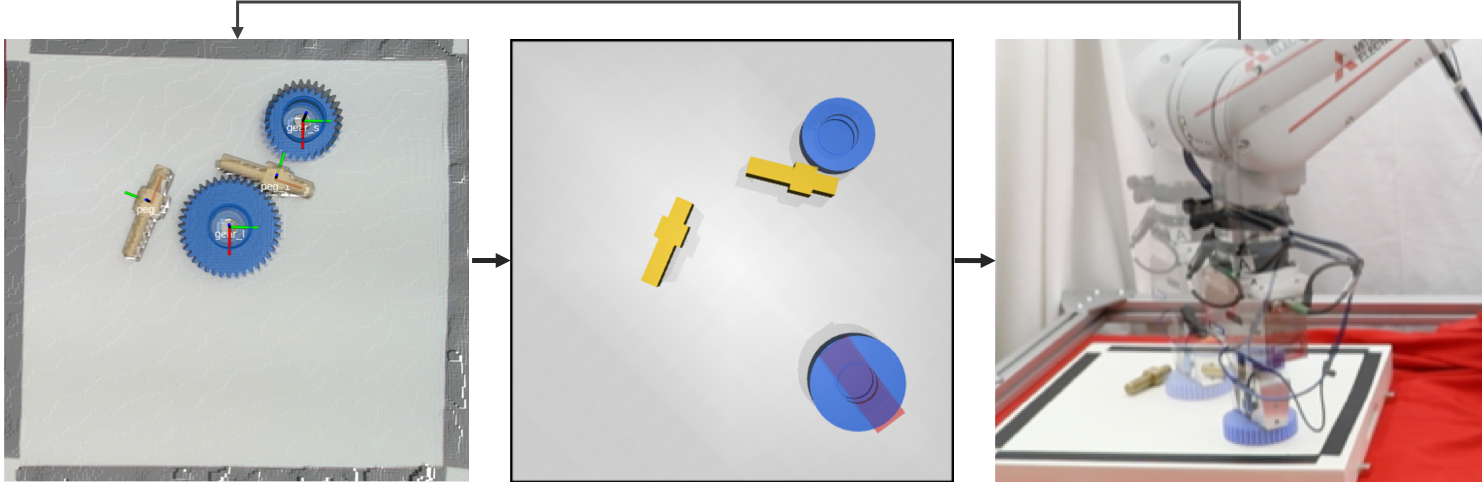}
    \caption{Singulation procedure. Given part pose from vision module (left), we first reconstruct the manipulation environment in the MuJoCo physics engine and generate an action using the random shooting method (middle). We then apply the action in the real system by sliding the target object using a suitable impedance controller with the F/T sensor (right).}
    \label{fig:singulation}
\end{minipage}
\begin{minipage}{0.07\linewidth}
\end{minipage}
\begin{minipage}{0.3\linewidth}
    \centering
    \includegraphics[width=0.9\columnwidth]{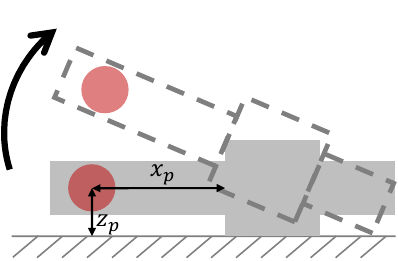}
    \caption{For performing the peg manipulation, we design a grasp that allows in-hand rotation as the peg is grasped from the table-top. The grasp is defined as $g_p=g_p(l_p,f_p)$, where $l_p=(x_p, z_p)$ as shown in the figure.}
    \label{fig:peg_manipulation}
\end{minipage}
\vspace{-0.2cm}
\end{figure*}

\subsection{Part Singulation} \label{sec:part_singulation}
The assembly parts are placed in any initial configuration in a clutter to allow generalization. Consequently, the robot first has to reason about grasp afforandance of the assembly parts. We design a model-based method using random-shooting to reason about part singulation. 

\textbf{Model-based Singulation using Random Shooting.}
Given the pose of all the objects from the vision system, we create the manipulation environment in simulation using the MuJoCo physics engine~\cite{todorov2012mujoco}. We use a simple random shooting method in this simulation environment to optimize the robot's action using a suitable cost function. More concretely, we sample a fixed number of actions and choose the one that minimizes the cost function. The action is described as $a = \{ o, \mathrm{d}x, \mathrm{d}y \}$, where $o \in \{ p_1,p_2,g_l, g_s \}$ is the object to interact with and $\mathrm{d}x, \mathrm{d}y$ are the relative planar positions on the manipulation surface to which the selected object $o$ is moved. We sample actions from a uniform distribution independently, specifically setting $\mathrm{d}x, \mathrm{d}y \sim U(-0.3, 0.3)$ [m] to sample a direction and a distance to move.

While pushing provides a good manipulation primitive to move the object on the table top surface, it is susceptible to uncertainty in friction and point of contact~\cite{SHIRAI2024101466}. To allow better robustness, we use the sliding manipulation primitive to move objects on the table-top~\cite{yi2023precise}. To imitate the sliding behavior in the simulation, we divide the direction $\mathrm{d}x, \mathrm{d}y$ into smaller steps and check the collision during the move of the object. We then compute the cost of the sampled action using the following function:
\begin{equation}
    c = -\lambda_1 \mathbb{I}_\mathrm{collision} + \lambda_2 \mathbb{I}_\mathrm{gear} - \lambda_3 d_\mathrm{obj} -\lambda_4 d_\mathrm{center}, \label{eq:random_shooting}
\end{equation}
where $\mathbb{I}_\mathrm{collision}$ and $\mathbb{I}_\mathrm{gear}$ are indicator functions of whether a collision between the selected object and other objects happens during sliding and whether a gear is selected. $d_\mathrm{obj}$ and $d_\mathrm{center}$ are the distance between the selected object and others before and after the sliding operation and the distance between the selected object from the center of the manipulation surface, respectively. The first term encourages optimization to avoid collisions, and the second one encourages the agent to interact with the gears since they have larger contact surfaces than pegs, resulting in fewer chances of losing contact during sliding on the real system. The third term encourages the selected object to have more affordance after the sliding operation. The final term avoids the part from dropping from the manipulation surface. We sample $N^\text{RS} = 100$ actions and choose the one with the minimum cost as the action to apply in the real system.

\textbf{Singulation execution by sliding.}
We do not consider the frictional interaction for sliding in simulation to simplify the sim2real transfer. Rather, we design a reliable and compliant sliding controller that can execute sliding for simple linear paths (see Fig.~\ref{fig:singulation} right). We use the object pose estimate from vision feedback to establish contact with the chosen object and then slide the object using a predefined normal contact force. The normal force is optimized on the real system to minimize slipping between the object and the manipulator end-effector. We terminate the singulation process if all pegs can be grasped without colliding with other objects. This is checked in the physics engine based on the vision feedback after every sliding action performed by the robot.

\subsection{Peg Grasping and Manipulation}
Once the parts are singulated, the objective is to grasp the singulated parts and perform the downstream assembly task. In order to assemble the gear box, the peg has to be inserted in the correct hole. Thus, apart from grasping the peg, it has to be re-oriented so that it could be inserted (see the third column of Fig.~\ref{fig:intro_fig}). There are several possibilities one can choose from. There is a wealth of work on using non-prehensile manipulation using extrinsic contacts that could be used to re-orient the peg~\cite{zhang2023simultaneous, 9811812, shirai2023robust}. However, non-prehensile manipulation is generally difficult to track online as one would need to estimate the slip, which requires a more careful design of tactile sensors and fingers. 

The other option we investigate is in-hand manipulation~\cite{liang2024robust, chavan2020planar}, as it could be tracked easily using tactile sensors co-located at the gripper fingers. We search for a grasp such that the peg re-orients in grasp as the peg is lifted from the table-top. In particular, we find a parameterized grasp denoted as $g_p=g_p(l_p,f_p)$, where $l_p$ is the grasp location in the local peg frame and $f_p$ is the grasping force that results in the desired peg re-orientation. This is shown in Fig.~\ref{fig:peg_manipulation}. 


\subsection{In-hand Pose Estimation}
\begin{figure}
    \centering
    \includegraphics[width=0.7\columnwidth]{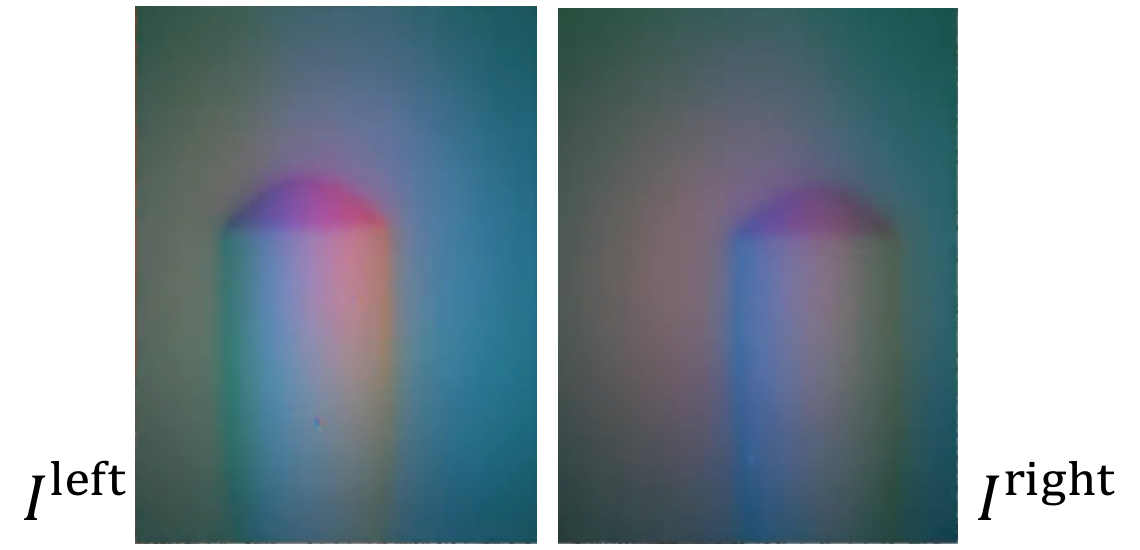}
    \caption{Example tactile images when grasping a peg, which is inputted to the in-hand pose estimation model to estimate the grasp error and correct it before insertion.}
    \label{fig:in_hand_pose_est}
    \vspace{-3mm}
\end{figure}

One of the biggest challenges with assembly is to localize the hole w.r.t. the peg after the robot has been able to grasp the peg. This presents a big challenge as with variation in grasp, achieving very tight localization between the peg and the hole could be challenging~\cite{9561646}. Previous approaches to performing tight tolerance insertion have tried to correct relative pose using force or tactile observations during contact formation with the environment (see~\cite{9561646,10178229,ota2024tactile}). However, allowing grasp to be a variable for the underlying problem makes learning challenging. An alternative approach could be to localize the peg in the grasp using a reliable in-hand pose estimation method. Vision-based tactile sensors co-located at the gripper fingers are very convenient for this task and have previously been explored for similar tasks~\cite{ota2023tactile,bauza2023tac2pose}. We train an in-hand pose estimation model that can predict the relative pose of the peg w.r.t. the local gripper frame. The model takes two tactile images obtained from the tactile sensors attached to the two fingers $I^\mathrm{left}, I^\mathrm{right}$ (see Fig.~\ref{fig:in_hand_pose_est}), and estimates the pose transformation in the local gripper frame. 


To train the model, we collect tactile images of the peg in the grasp by introducing displacement from a uniform distribution of $\mathrm{d}x \sim U(-10, 10)$ [mm] in the local gripper frame. We then train a ResNet18~\cite{he2016deep} model with regression loss to predict the displacement in the local gripper frame.
Since the hole position is known in the local gripper frame, we can perform insertion once the peg is localized in the local gripper frame.


\subsection{Gear Insertion}
After peg insertion, the robot needs to insert the gears to assemble the gear box. Similar to the peg insertion problem, grasping the gears using vision-based localization introduces grasp uncertainty, and thus the robot needs to localize the gear in the local grasp (or gripper) frame. Even though the tactile sensors provide us with very high-resolution object features, tactile images do not provide enough information for in-hand localization of gears. This is because the gear teeth are identical and do not have any additional features to estimate the pose of the gear due to their limited field of view. Consequently, we make use of the force observations up on contact formation between the gear and the peg to localize the gear w.r.t. the peg. This is similar to the method presented earlier in~\cite{10178229}. Although one can train such a policy using tactile sensors (similar to~\cite{9561646}), we make use of six-axis force observations from the F / T sensor mounted on the robot wrist. 


To train the gear insertion policy, we collect data by introducing known error between the gear center and the peg. We then measure the six-axis force observations up on contact formation between the peg and the gear with known pose error. The sensor can return measurements at $280$ Hz. To collect training data, we add the known pose error between the gear and the peg, such as $\mathrm{d}x \sim \{ -4, 4 \}$ [mm]. The robot then makes contact with the peg, and we collect data for $T=3$ seconds for each insertion attempt. After collecting the raw signals, we compute the moving average over $70$ time steps, since the raw F/T signals are noisy. 

\subsection{Gear Meshing}
\begin{figure}
    \centering
    \includegraphics[width=0.8\columnwidth]{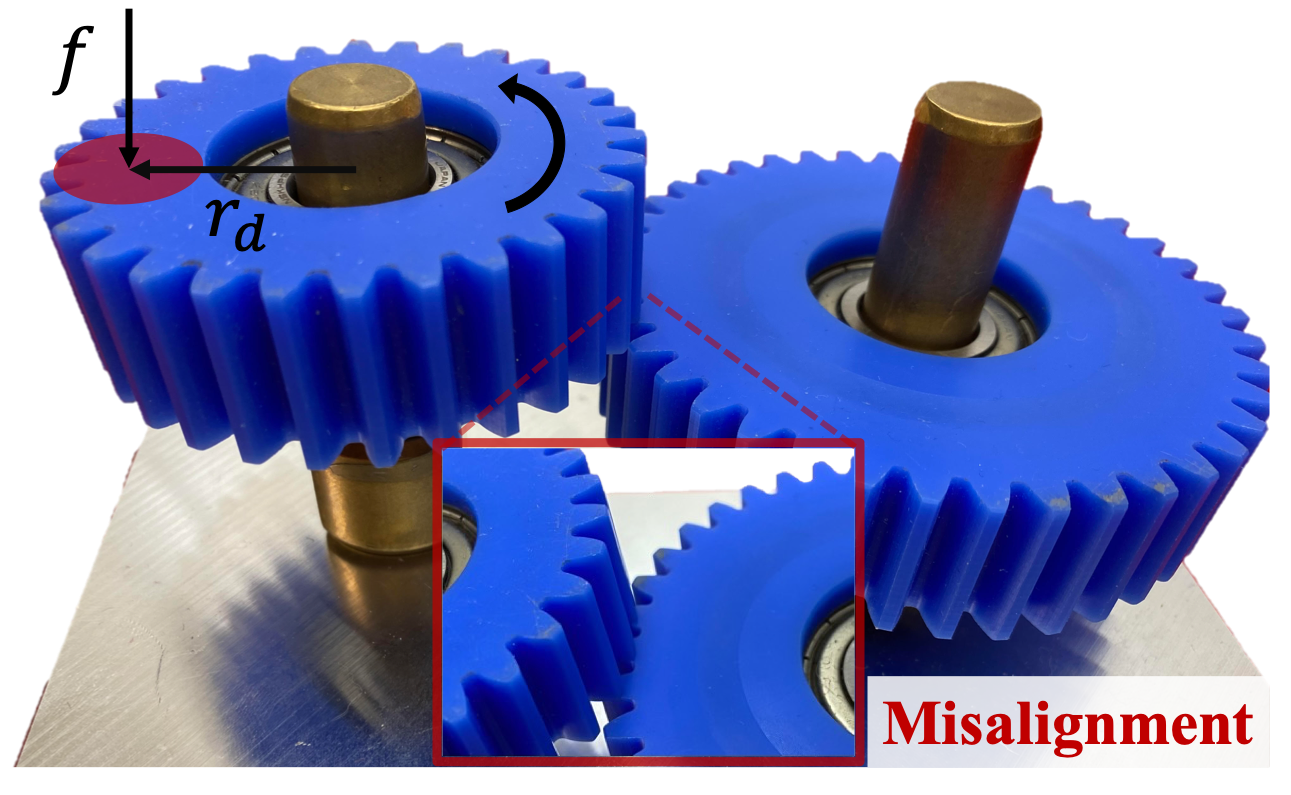}
    \caption{This figure shows the misalignment between the gears during meshing and the controller designed to align the gear teeth by using a suitable compliance control input $u=u(f,r_d)$. This compliance controller is used to rotate the smaller gear while creating relative movement between the gears so that eventually the teeth align.}
    \label{fig:gear_meshing}
    \vspace{-8pt}
\end{figure}

The tolerances of the parts are such that simply inserting the gears does not result in alignment of the gear teeth (see Fig.~\ref{fig:gear_meshing}). Consequently, the robot has to perform fine manipulation so that the gear teeth could be meshed. We first investigated a force-feedback strategy to understand the alignment of the gear teeth. However, we observed that there was no reliable trend between the observed force signature and teeth alignment. Thus, we propose an open-loop compliant gear meshing method. In particular, we designed a compliant force controller $u=u(f,r_d)$ where $f$ is the meshing force and $r_d$ is the radial distance from the gear center at which the force is applied, as shown in Fig.~\ref{fig:gear_meshing}. This compliant controller has the task of rotating the small gear while minimally disturbing the large gear. Note that the large gear will still rotate due to friction between the two gears. However, the compliant controller simply needs to create relative movement between the two gears. Due to the relative movement between the gears, the gear teeth eventually align.

\section{Results}\label{sec:results}
We perform rigorous experiments to understand and address the following questions.
\begin{itemize}
    \item What is the success rate of the individual system components? What are some of the observed failure cases?
    \item What is the success rate and performance of the overall system?
\end{itemize}
In the rest of the paper, we present experiments to address these questions. We skip detailed analysis of compliance controller for peg manipulation and gear meshing for brevity and will be presented in an extended version of the paper.

\subsection{Vision Pose Estimation}
\begin{figure}
    \centering
    \includegraphics[width=1.0\columnwidth]{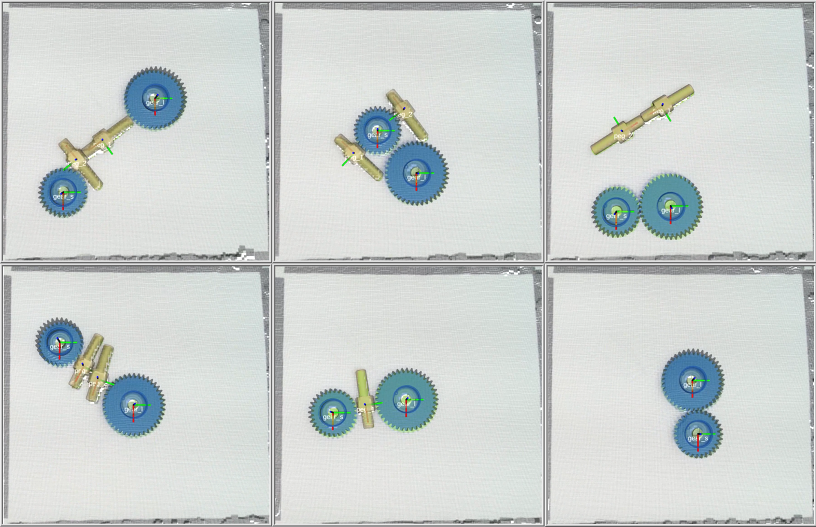}
    \caption{\textbf{Vision Pose Estimation.} Qualitative results from the vision system. CAD model overlayed on point cloud for pose visualization. }
    \label{fig:vision_qual}
    \vspace{-6pt}
\end{figure}

We report detection and instance segmentation results for validation (Table~\ref{tab:vision_quant}). We use COCO evaluation metrics AP (averaged over IoU thresholds). 
Figure~\ref{fig:vision_qual} shows a sample of qualitative results from the vision system. The efficacy of vision-based pose estimation is assessed on downstream tasks. Our findings indicate consistent success in effectively grasping and singulating assembly parts from vision across diverse test scenarios. The final precision required for insertion tasks was compensated by tactile and F/T sensor feedback, which is presented next.

\begin{table}[t]
    \centering
    \caption{\textbf{Vision Pose Estimation.} Validation results for both detection and instance segmentation. APb and APm denotes detection and instance segmentation results.}
    \begin{tabular}{cccccc}
        \toprule
        APb & APb@0.5 & APb@0.75 & APm & APm@0.5 & APm@0.75 \\
        \midrule
        93.32 & 99.50 & 99.08 & 93.37 & 99.50 & 99.08 \\ 
        \bottomrule
    \end{tabular}
    \label{tab:vision_quant}
\end{table}

\begin{table}[t]
    \centering
    \caption{\textbf{Part singulation.} Success rate, number of interactions, and elapsed time required to singulate objects with different number of samples $N^\text{RS}$ in simulation and the real system. The results are averaged over $100$ runs on different initial conditions.}
    \begin{tabular}{r|cccc|c} \toprule
                           & \multicolumn{4}{c}{Simulation} & Real\\
    Number of samples $N^\text{RS}$ &  $1$ & $10$ & $100$ & $1000$ & $100$ \\ \midrule
    Success rate           & $0.19$ & $0.80$ & $1.00$ & $1.00$ & $1.00$ \\
    Number of interactions & $8.78$ & $4.12$ & $1.94$ & $1.94$ & $3.26$ \\
    Elapsed time [secs]    & $0.05$ & $0.21$ & $1.38$ & $12.87$ & -\\ \bottomrule
    \end{tabular}
    \label{tab:singulation}
\end{table}

\subsection{Part Singulation}
To evaluate the performance of part singulation, we compare the random shooting method with different numbers of samples, $N^\text{RS} \in \{1, 10, 100, 1000 \}$. We set the maximum number of interactions to $10$, and use this value if the algorithm fails to singulate, that is, the target object moves outside the manipulation surface. 
Table.~\ref{tab:singulation} shows the success rate, number of interactions, and elapsed time to complete the singulation. As expected, the success rate and the number of interactions improve as the number of samples increases and converge with $N^\text{RS}=100$. Since $N^\text{RS}=1000$ takes approximately $9.3$ times longer than $N^\text{RS}=100$ with similar performance, we use $N^\text{RS}=100$ for real experiments.

\begin{figure}
    \centering
    \includegraphics[width=0.85\columnwidth]{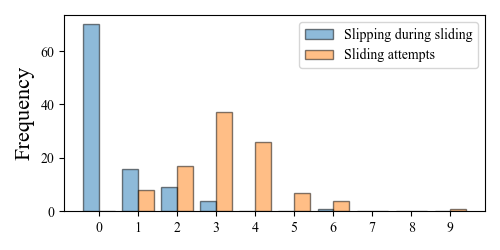}
    \caption{\textbf{Part singulation.} The histogram shows the distribution of singulation attempts and slip frequency between the object and the manipulator over 100 real-system trials.
    }
    \label{fig:singulation_histogram}
    \vspace{-4pt}
\end{figure}

The model-based planner for singulation was implemented on the real system using a compliance controller designed for sliding. We tested the execution on real systems for $100$ initial conditions with a $100\%$ success rate. The number of interactions required for singulation is presented in Fig.~\ref{fig:singulation_histogram}. We also observed slipping between the manipulator and objects during sliding, as shown in the histogram.

\subsection{In-hand Pose Estimation}
Next, we evaluate the performance of the in-hand pose estimation. We compare our method with a different number of training data ($10, 100, 1000$) to a pick-and-place policy, where the robot does not correct the grasping error and tries to insert the peg directly into the hole. To test the methods, we randomly added an error in grasp on the $X$-axis from a uniform distribution $dx \sim U(-10, 10)$ [mm].

The results in Table.~\ref{tab:insertion} demonstrate that the baseline pick-and-place policy fails to insert the peg because the tolerance of the target hole is very tight ($1$ mm). On the contrary, our method with sufficient training data (more than 100) achieves a submillimeter precision in estimation, and enables the robot to insert the peg successfully. Data collection time for $100$ data requires approximately $20$ minutes.

\begin{table}[t]
    \centering
    \caption{\textbf{In-hand pose estimation.} success rate and average error of baseline and our method with different number of training data $(10, 100, 1000)$.}
    \begin{tabular}{r|cccc} \toprule
                 & \multirow{2}{*}{Baseline} & \multicolumn{3}{c}{Ours} \\
                 &                           & $10$ & $100$ & $1000$  \\\midrule
        Success rate  & $3/20$ & $10/20$ & $20/20$ & $20/20$ \\ 
        Average error [mm] & N/A & $1.52$ & $0.43$ & $0.18$ \\ \bottomrule \end{tabular} \label{tab:insertion}
\end{table}

\subsection{Gear Insertion}
\begin{figure}[t]
    \centering
    \includegraphics[width=\columnwidth]{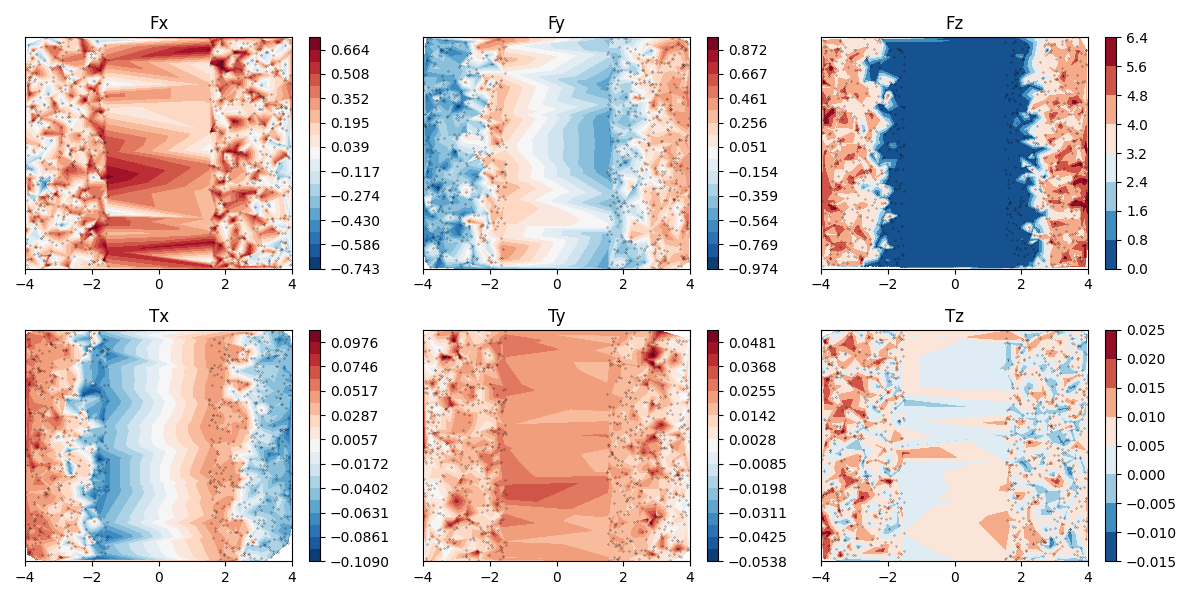}
    \caption{\textbf{Gear insertion.} Force observations up on contact formation between the large gear and peg. We use these to learn the dependence between the alignment and the expected force readings. The horizontal axis is the error in $X$-axis.}
    \label{fig:gear_insertion}
    \includegraphics[width=0.85\columnwidth]{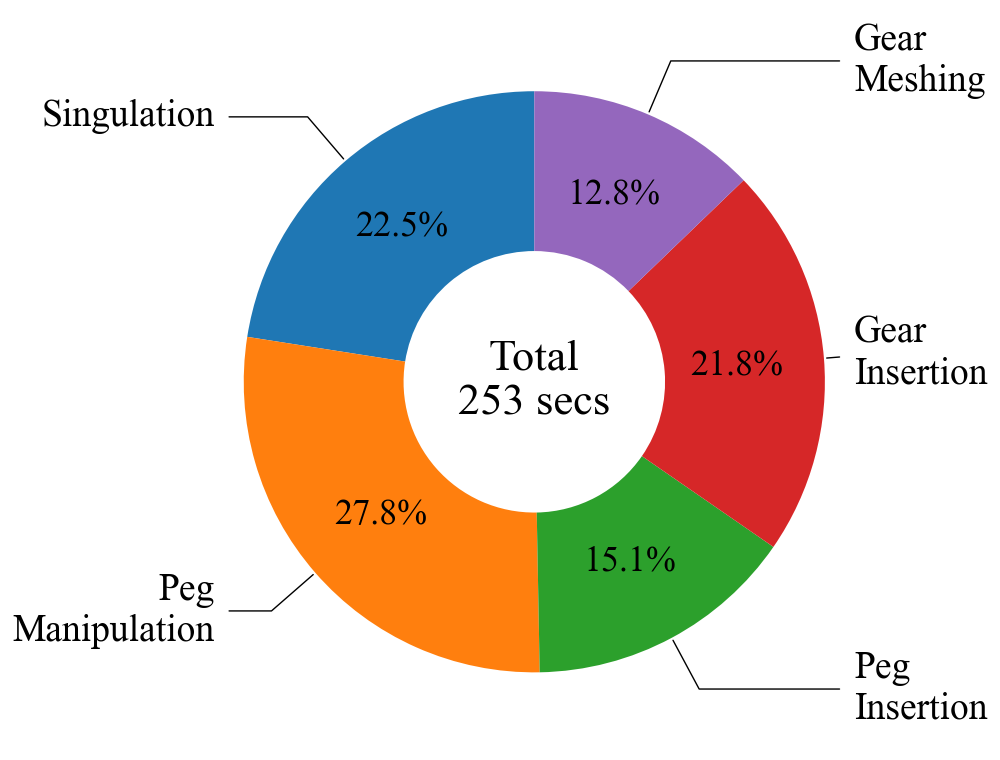}
    \caption{\textbf{End-to-End evaluation.} Elapsed time for each operation averaged over $20$ runs.}
    \label{fig:elapsed_time}
\vspace{-0.6cm}
\end{figure}

Figure~\ref{fig:gear_insertion} shows the variation of the force data against the relative error between the peg and gear position during insertion. This data is then used to train a supervised gear insertion policy that can predict the expected error based on the force readings. We use LSTM~\cite{hochreiter1997long} with two layers, each having $256$ units, to capture patterns in time-series data. The supervised gear insertion predicts the direction of error between the peg and the gear in an iterative fashion. This supervised policy was tested for $100$ times where it was able to achieve a success rate of $95\%$ for gear insertion. 


\subsection{End-to-End Evaluation}
We have tested the entire system under $225$ different initial conditions (randomized by a diverse set of audience), and have been able to achieve a success rate of $99.11$ \% ($223/225)$. In one of the failure cases, the robot could not singulate the parts. The other failure case was when the robot could not mesh the gear after $5$ attempts. Fig.~\ref{fig:elapsed_time} shows the average elapsed time for each operation over $20$ different runs. The majority of time is dedicated to manipulating the peg, demanding delicate adjustments and re-grasping (see supplementary video). Misalignment related to gear grasping necessitate intervention from the gear insertion policy trained on force signals. This tendency is evident in the duration of gear insertion. Notably, the timing results encompass the manipulator arm's movement time.

\section{Conclusion and Discussion}\label{sec:conclusion}
Creating autonomous assembly agents can have a huge impact on the manufacturing industry. However, designing such agents presents myriads of challenges to achieve operational flexibility and robustness. In order to understand some of the challenges associated with autonomous assembly, we presented an autonomous robotic assembly system. The proposed assembly system can perform singulation and pose manipulation of assembly parts followed by high-precision assembly using vision as well as tactile feedback. The proposed system is extensively tested with various different initial configurations of the assembly parts to demonstrate the robustness of the proposed system. 

We discovered several different failure cases during the development of the assembly system, which gave us meaningful useful insights. Since the singulation planning used a random shooting method, oftentimes it resulted in sub-optimal plans. The in-hand manipulation of the peg would fail after continued use of the Gelsight sensors. This was mainly because the gel pads of the sensors would become smooth, thus resulting in a reduction of the friction coefficient causing an undesirable slip of the peg during the in-hand manipulation. Similarly, the gear meshing was very susceptible to the amount of force applied during the meshing operation, requiring very fine manipulation. In the future, it would be desirable to have the capability of detecting these failures automatically and to have failure recovery modules. 

Although we demonstrate the effectiveness of the proposed system with familiar components, there's a crucial need for robots to handle unfamiliar parts. Achieving this entails enabling robots to swiftly perceive and construct object models through interactive perception. This process mirrors how humans intuitively form object models during interactions. Nevertheless, comprehensively understanding the creation of object models tailored for object manipulation remains an area ripe for exploration. 

\bibliographystyle{IEEEtran}

\bibliography{references}

\begin{thebibliography}{10}
\providecommand{\url}[1]{#1}
\csname url@rmstyle\endcsname
\providecommand{\newblock}{\relax}
\providecommand{\bibinfo}[2]{#2}
\providecommand\BIBentrySTDinterwordspacing{\spaceskip=0pt\relax}
\providecommand\BIBentryALTinterwordstretchfactor{4}
\providecommand\BIBentryALTinterwordspacing{\spaceskip=\fontdimen2\font plus
\BIBentryALTinterwordstretchfactor\fontdimen3\font minus \fontdimen4\font\relax}
\providecommand\BIBforeignlanguage[2]{{%
\expandafter\ifx\csname l@#1\endcsname\relax
\typeout{** WARNING: IEEEtran.bst: No hyphenation pattern has been}%
\typeout{** loaded for the language `#1'. Using the pattern for}%
\typeout{** the default language instead.}%
\else
\language=\csname l@#1\endcsname
\fi
#2}}

\bibitem{zhu2023multi}
X.~Zhu, D.~K. Jha, D.~Romeres, L.~Sun, M.~Tomizuka, and A.~Cherian, ``Multi-level reasoning for robotic assembly: From sequence inference to contact selection,'' \emph{arXiv preprint arXiv:2312.10571}, 2023.

\bibitem{mason2018toward}
M.~T. Mason, ``Toward robotic manipulation,'' \emph{Annual Review of Control, Robotics, and Autonomous Systems}, vol.~1, pp. 1--28, 2018.

\bibitem{knepper2013ikeabot}
R.~A. Knepper, T.~Layton, J.~Romanishin, and D.~Rus, ``Ikeabot: An autonomous multi-robot coordinated furniture assembly system,'' in \emph{Proc. of ICRA}.\hskip 1em plus 0.5em minus 0.4em\relax IEEE, 2013, pp. 855--862.

\bibitem{thomas2018learning}
G.~Thomas, M.~Chien, A.~Tamar, J.~A. Ojea, and P.~Abbeel, ``Learning robotic assembly from cad,'' in \emph{Proc. of ICRA}, 2018, pp. 3524--3531.

\bibitem{suarez2018can}
F.~Su{\'a}rez-Ruiz, X.~Zhou, and Q.-C. Pham, ``Can robots assemble an ikea chair?'' \emph{Science Robotics}, vol.~3, no.~17, p. eaat6385, 2018.

\bibitem{kimbel2020benchmarking}
K.~Kimble, K.~Van~Wyk, J.~Falco, E.~Messina, Y.~Sun, M.~Shibata, W.~Uemura, and Y.~Yokokohji, ``Benchmarking protocols for evaluating small parts robotic assembly systems,'' \emph{IEEE Robotics and Automation Letters}, vol.~5, no.~2, pp. 883--889, 2020.

\bibitem{mahler2019learning}
J.~Mahler, M.~Matl, V.~Satish, M.~Danielczuk, B.~DeRose, S.~McKinley, and K.~Goldberg, ``Learning ambidextrous robot grasping policies,'' \emph{Science Robotics}, vol.~4, no.~26, p. eaau4984, 2019.

\bibitem{kiatos2019robust}
M.~Kiatos and S.~Malassiotis, ``Robust object grasping in clutter via singulation,'' in \emph{Proc. of ICRA}.\hskip 1em plus 0.5em minus 0.4em\relax IEEE, 2019, pp. 1596--1600.

\bibitem{lundell2021ddgc}
J.~Lundell, F.~Verdoja, and V.~Kyrki, ``Ddgc: Generative deep dexterous grasping in clutter,'' \emph{IEEE Robotics and Automation Letters}, vol.~6, no.~4, pp. 6899--6906, 2021.

\bibitem{zeng2018learning}
A.~Zeng, S.~Song, S.~Welker, J.~Lee, A.~Rodriguez, and T.~Funkhouser, ``Learning synergies between pushing and grasping with self-supervised deep reinforcement learning,'' in \emph{Proc. of IROS}.\hskip 1em plus 0.5em minus 0.4em\relax IEEE, 2018, pp. 4238--4245.

\bibitem{xiang2017posecnn}
Y.~Xiang, T.~Schmidt, V.~Narayanan, and D.~Fox, ``Posecnn: A convolutional neural network for 6d object pose estimation in cluttered scenes,'' \emph{RSS}, 2018.

\bibitem{tremblay2018deep}
J.~Tremblay, T.~To, B.~Sundaralingam, Y.~Xiang, D.~Fox, and S.~Birchfield, ``Deep object pose estimation for semantic robotic grasping of household objects,'' \emph{CoRL}, 2018.

\bibitem{bauza2023tac2pose}
M.~Bauza, A.~Bronars, and A.~Rodriguez, ``Tac2pose: Tactile object pose estimation from the first touch,'' \emph{The International Journal of Robotics Research}, vol.~42, no.~13, pp. 1185--1209, 2023.

\bibitem{ota2023tactile}
K.~Ota, D.~K. Jha, H.-Y. Tung, and J.~B. Tenenbaum, ``Tactile-filter: Interactive tactile perception for part mating,'' \emph{RSS}, 2023.

\bibitem{9561646}
S.~Dong, D.~K. Jha, D.~Romeres, S.~Kim, D.~Nikovski, and A.~Rodriguez, ``Tactile-rl for insertion: Generalization to objects of unknown geometry,'' in \emph{Proc. of ICRA}, 2021, pp. 6437--6443.

\bibitem{zhang2023simultaneous}
M.~Zhang, D.~K. Jha, A.~U. Raghunathan, and K.~Hauser, ``Simultaneous trajectory optimization and contact selection for multi-modal manipulation planning,'' \emph{RSS}, 2023.

\bibitem{9811812}
Y.~Shirai, D.~K. Jha, A.~U. Raghunathan, and D.~Romeres, ``Robust pivoting: Exploiting frictional stability using bilevel optimization,'' in \emph{Proc. of ICRA}, 2022, pp. 992--998.

\bibitem{shirai2023robust}
Y.~Shirai, D.~K. Jha, and A.~U. Raghunathan, ``Robust pivoting manipulation using contact implicit bilevel optimization,'' \emph{arXiv preprint arXiv:2303.08965}, 2023.

\bibitem{chavan2020planar}
N.~Chavan-Dafle, R.~Holladay, and A.~Rodriguez, ``Planar in-hand manipulation via motion cones,'' \emph{The International Journal of Robotics Research}, vol.~39, no. 2-3, pp. 163--182, 2020.

\bibitem{kim2024texterity}
S.~Kim, A.~Bronars, P.~Patre, and A.~Rodriguez, ``Texterity--tactile extrinsic dexterity: Simultaneous tactile estimation and control for extrinsic dexterity,'' \emph{arXiv preprint arXiv:2403.00049}, 2024.

\bibitem{liang2024robust}
B.~Liang, K.~Ota, M.~Tomizuka, and D.~Jha, ``Robust in-hand manipulation with extrinsic contacts,'' \emph{arXiv preprint arXiv:2403.18960}, 2024.

\bibitem{10178229}
D.~K. Jha, D.~Romeres, S.~Jain, W.~Yerazunis, and D.~Nikovski, ``Design of adaptive compliance controllers for safe robotic assembly,'' in \emph{2023 European Control Conference (ECC)}, 2023, pp. 1--8.

\bibitem{heo2023furniturebench}
M.~Heo, Y.~Lee, D.~Lee, and J.~J. Lim, ``Furniturebench: Reproducible real-world benchmark for long-horizon complex manipulation,'' in \emph{RSS}, 2023.

\bibitem{lee2021rgbstacking}
A.~X. Lee and et~al., ``Beyond pick-and-place: Tackling robotic stacking of diverse shapes,'' in \emph{CoRL}, 2021.

\bibitem{bousmalis2024robocat}
K.~Bousmalis, G.~Vezzani, D.~Rao, C.~M. Devin, A.~X. Lee, and et~al., ``Robocat: A self-improving generalist agent for robotic manipulation,'' \emph{Transactions on Machine Learning Research}, 2024.

\bibitem{bauza2023simple}
M.~Bauza, A.~Bronars, Y.~Hou, I.~Taylor, N.~Chavan-Dafle, and A.~Rodriguez, ``simple: a visuotactile method learned in simulation to precisely pick, localize, regrasp, and place objects,'' \emph{arXiv preprint arXiv:2307.13133}, 2023.

\bibitem{sun2023interactive}
L.~Sun, D.~K. Jha, C.~Hori, S.~Jain, R.~Corcodel, X.~Zhu, M.~Tomizuka, and D.~Romeres, ``Interactive planning using large language models for partially observable robotics tasks,'' \emph{Proc. of ICRA}, 2024.

\bibitem{gelsightmini2023}
``{GelSight Mini},'' \url{https://www.gelsight.com/gelsightmini/}, accessed: 2023-01-16.

\bibitem{wen2023bundlesdf}
B.~Wen, J.~Tremblay, V.~Blukis, S.~Tyree, T.~M{\"u}ller, A.~Evans, D.~Fox, J.~Kautz, and S.~Birchfield, ``Bundlesdf: Neural 6-dof tracking and 3d reconstruction of unknown objects,'' in \emph{Proc. of CVPR}, 2023.

\bibitem{he2016deep}
K.~He, X.~Zhang, S.~Ren, and J.~Sun, ``Deep residual learning for image recognition,'' in \emph{Proc. of CVPR}, 2016, pp. 770--778.

\bibitem{he2017mask}
K.~He, G.~Gkioxari, P.~Doll{\'a}r, and R.~Girshick, ``Mask r-cnn,'' in \emph{Proc. of CVPR}, 2017, pp. 2961--2969.

\bibitem{jain2016grasp}
S.~Jain and B.~Argall, ``Grasp detection for assistive robotic manipulation,'' in \emph{Proc. of ICRA}.\hskip 1em plus 0.5em minus 0.4em\relax IEEE, 2016, pp. 2015--2021.

\bibitem{todorov2012mujoco}
E.~Todorov, T.~Erez, and Y.~Tassa, ``Mujoco: A physics engine for model-based control,'' in \emph{Proc. of IROS}, 2012, pp. 5026--5033.

\bibitem{SHIRAI2024101466}
Y.~Shirai, D.~K. Jha, A.~U. Raghunathan, and D.~Romeres, ``Chance-constrained optimization for contact-rich systems using mixed integer programming,'' \emph{Nonlinear Analysis: Hybrid Systems}, vol.~52, p. 101466, 2024.

\bibitem{yi2023precise}
X.~Yi and N.~Fazeli, ``Precise object sliding with top contact via asymmetric dual limit surfaces,'' \emph{RSS}, 2023.

\bibitem{ota2024tactile}
K.~Ota, D.~K. Jha, K.~M. Jatavallabhula, A.~Kanezaki, and J.~B. Tenenbaum, ``Tactile estimation of extrinsic contact patch for stable placement,'' in \emph{Proc. of ICRA}, 2024.

\bibitem{hochreiter1997long}
S.~Hochreiter and J.~Schmidhuber, ``Long short-term memory,'' \emph{Neural computation}, vol.~9, no.~8, pp. 1735--1780, 1997.

\end{thebibliography}

\end{document}